\documentclass[twoside,leqno,twocolumn]{article}

\usepackage[letterpaper]{geometry}

\usepackage{style_files/ltexpprt}
\usepackage{graphicx} 
\usepackage{caption} 
\usepackage{algorithm}
\usepackage{algorithmic}
\usepackage{newfloat}
\usepackage{listings}
\usepackage{times}

\usepackage{soul}
\usepackage{url}
\usepackage[hidelinks]{hyperref}
\usepackage[utf8]{inputenc}
\usepackage{amsmath}
\usepackage{booktabs}
\usepackage{refcount}
\usepackage{amsfonts}
\usepackage{caption}
\usepackage{color, colortbl}
\usepackage{wrapfig}
\usepackage{breqn}
\usepackage{bm}
\usepackage{enumerate}
\usepackage{subfigure}
\usepackage{subfloat}
\usepackage{multirow}
\usepackage{threeparttable}
\usepackage[table,xcdraw]{xcolor}

\newtheorem{definition}{Definition}
\newtheorem{pro-stat}{Problem Definition}
\def\T{{\scriptscriptstyle\mathsf{T}}}

\DeclareMathOperator*{\pooling}{Pooling}

\newcommand{\nemo}{\textsc{NeMo}}
\definecolor{Gray}{gray}{0.80}
\definecolor{LightGray}{gray}{0.93}

\newcommand{\hh}[1]{{\small\color{red}{\bf HH: #1}}}
\newcommand{\bx}[1]{{\small\color{blue}{\bf BX: #1}}}

\newcommand{\hide}[1]{}
\newcommand{\mkclean}{
	\renewcommand{\hh}[1]{}
	\renewcommand{\bx}[1]{}
}

\DeclareCaptionStyle{ruled}{labelfont=normalfont,labelsep=colon,strut=off} 
\floatstyle{ruled}
\newfloat{listing}{tb}{lst}{}
\floatname{listing}{Listing}
\title{Neural Multi-network Diffusion towards Social Recommendation}
\author{
    Boxin Du*, Lihui Liu*, Jiejun Xu$\dagger$, Fei Wang$\ddagger$, Hanghang Tong* \\
    *Department of Computer Science, University of Illinois Urbana-Champaign\\
    $\dagger$ HRL, jxu@hrl.com\\
    $\ddagger$ Amazon, feiww@amazon.com \\
    *\{boxindu2, lihuil2, htong\}@illinois.edu 
}

\begin{document}
\setlength{\abovedisplayskip}{1.4pt}
\setlength{\belowdisplayskip}{1.4pt}
\setlength{\abovedisplayshortskip}{1.4pt}
\setlength{\belowdisplayshortskip}{1.4pt}
\maketitle

\mkclean

\begin{abstract}
Graph Neural Networks (GNNs) have been widely applied on a variety of real-world applications, such as social recommendation. However, existing GNN-based models on social recommendation suffer from serious problems of generalization and oversmoothness, because of the underexplored negative sampling method and the direct implanting of the off-the-shelf GNN models. In this paper, we propose a succinct \hh{let's not call our method 'simple' or 'simplified', use a neutral or positive word instead, such as succinct} multi-network GNN-based neural model (\nemo) \hh{NeMO is  a great name, although LL has used it for his paper a while ago \url{https://dl.acm.org/doi/10.1145/3041021.3054200} -- let us use MeNo for submission and we can decide later if we can find an alternative name} for social recommendation. Compared with the existing methods, the proposed model explores a generative negative sampling strategy, and leverages both the positive and negative user-item interactions for users' interest propagation. 
The experiments show that \nemo\ outperforms the state-of-the-art baselines on various real-world benchmark datasets (e.g., by up to $38.8\%$ in terms of NDCG@15). \hh{mention some numbers here, and/or conclusion}
\end{abstract}

\section{Introduction}
\hide{
\bx{outline: Major limitations of current methods: 
\begin{itemize}
    \item Negative sampling method for social recommendation is underexplored.
    \item The GNN-based neural models for social recommendation often suffer from oversmoothing issue. 
    \item The GNN-based neural models do not fully utilize the user-item interaction for message aggregation and representation learning, since they only consider the observed interactions. 
\end{itemize}
In this paper, we propose a succinct multi-network GNN-based neural model (\nemo) for social recommendation. Compared with the existing methods, the proposed model bears the following distinctive advantages:
\begin{itemize}
    \item A novel generative negative sampling method, which generates hard negative samples as a complement of the sampled negative samples. 
    \item A succinct multi-network GNN-based model, which does not adopt the common GNN models in the user representation learning, but instead only keeping the feature aggregation. 
    \item Leverage both the positive and negative user-item interactions in the users' interest propagation to explicitly model users' positive and negative preferences. 
\end{itemize}}
}

Graph Neural Networks (GNNs) are powerful tools for a large variety of machine learning and data mining tasks on graph data, such as node/graph classification \cite{hamilton2017inductive} \cite{du2019mrmine}, link prediction \cite{zhang2018link}, graph matching \cite{li2019graph, du2017first, liu2019g}, and graph-based recommendation \cite{he2020lightgcn}. Particularly, social relations play an important role in influencing users' preferences in recommender systems. When choosing from numerous items, users tend to follow their social network friends with whom they share the same interest or whom they trust. This type of trust/interest influence could naturally be captured by GNN models. As a result, substantial recent works focus on applying GNN techniques on the social recommendation task, in which the GNN model is leveraged for learning representations of users and items in a convolutional manner on social networks.

Despite remarkable successes, the existing works exhibit a few non-ignorable limitations. First, similar to traditional recommender systems, the negative sampling method for social recommendation is underexplored. One of the most common strategies is uniform negative sampling (UNS), in which samples from the unobserved items as negative samples with equal probability. However, the naive UNS method could introduce bias to the model since the unobserved items might also contain positive items. Second, the existing GNN-based neural models often directly adopt the off-the-shelf GNN models/layers, which often suffer from oversmoothing, especially for the social recommendation task. Furthermore, many current GNN architectures are tailored for specific tasks, and might not be suitable for social recommendation if applied directly. Third, the existing social recommendation models do not fully utilize the use-item interaction for message aggregation and representation learning, since they only consider the observed interactions for interest diffusion. Whereas the unobserved interactions and the negative samples might provide a different type of interactions. 

In this paper, we propose a multi-network GNN-based neural model (\nemo) for social recommendation problem. Compared with the existing methods, the proposed model bears the following distinctive advantages. First, we propose a novel generative negative sampling method, which aims at generating hard negative samples as a complement of the sampled negative samples to improve the generalization ability of the model. Second, we propose a succinct multi-network GNN-based model, which does not adopt existing sophisticated GNN models in the user representation learning, but instead selectively keeps a limited number of feature aggregation processes in order to mitigate the oversmoothing issue. Third, we leverage both the positive and negative user-item interactions for the users' interest propagation between users and items, in order to explicitly model the users' positive and negative preferences when learning the representations of users and items. 

The rest of the paper is organized as follows. In Section \ref{sec:problem}, we describe and formally define the social recommendation problem with preliminaries. In Section \ref{sec:method}, we elaborate the model architecture of \nemo\ with the proposed generative negative sampling strategy. In Section \ref{sec:experiments}, we present the empirical evaluation results. In Section \ref{sec:related-work}, we review the recent works. The paper is concluded in Section \ref{sec:conclusion}. 


\section{Problem Definition and Preliminaries}
\label{sec:problem}
In this section, we formally define the graph-based social recommendation problem and provide some preliminaries. 

\subsection{Problem Definition}
\begin{figure}
    \centering
    \includegraphics[width=0.47\textwidth, height=0.3\textwidth]{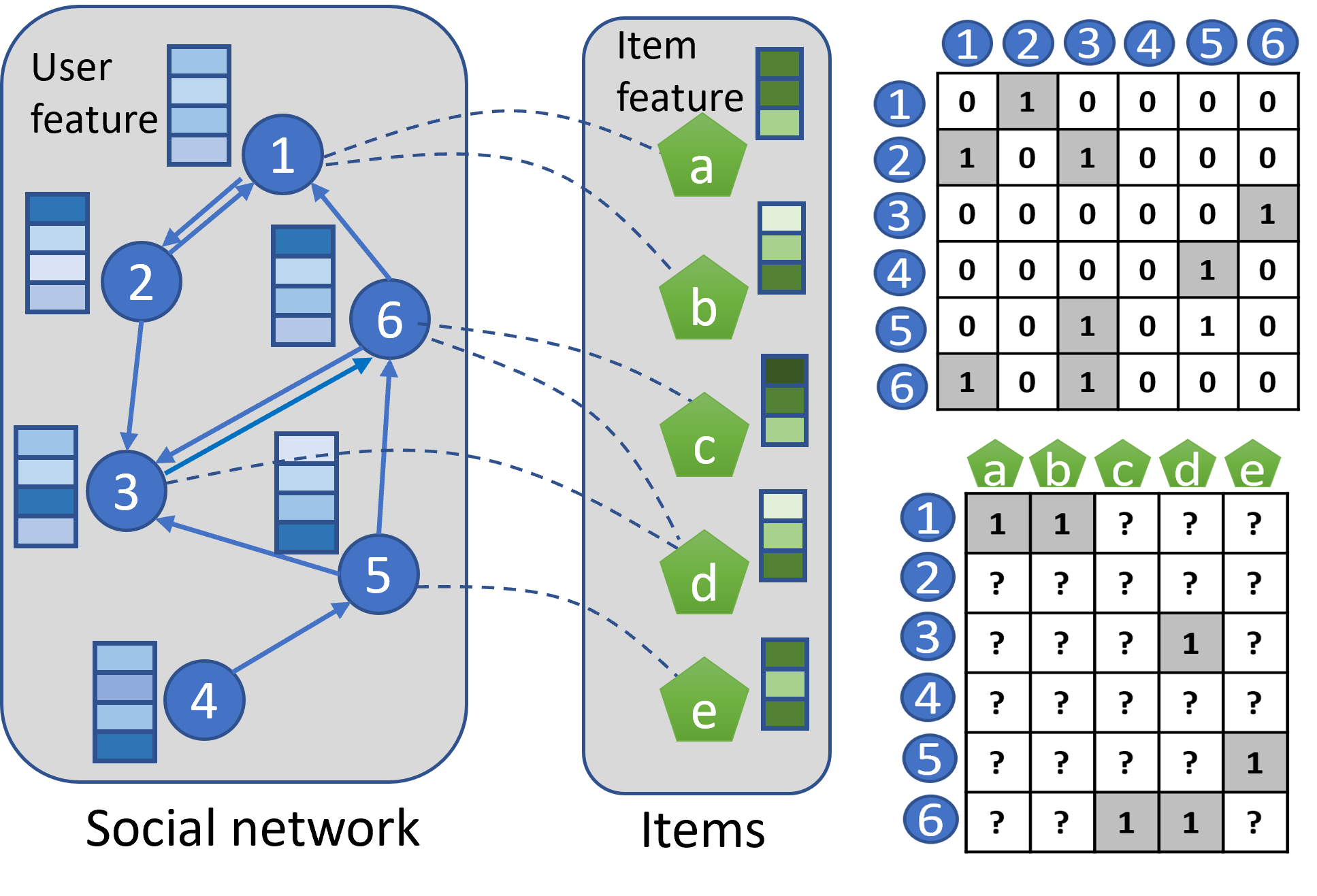}
    \caption{Social recommendation problem setting.}
    \label{fig:social_rec}
\end{figure}

In the traditional recommender system with users' implicit feedbacks, given a user set $\mathcal{U} = \{u_1, u_2, ..., u_N\}$ and item set $\mathcal{I} = \{i_1, i_2, ..., i_M\}$, there are only user-item interactions, which could be represented as a matrix $\mathbf{R} \in \mathbb{R}^{N\times M}$. $\mathbf{R}_{ui}=1$ if there is an observed interaction between user $u$ and item $i$, and otherwise unknown (denoted as $?$ in Figure \ref{fig:social_rec}). In the general social recommender setting, there is also a social network that reflects the relations and interactions among the users. Such networks often come with numerical feature attributes associated with both the users and items. The social relations can naturally be represented as an adjacency matrix $\mathbf{A} \in \mathbb{R}^{N \times N}$. We denote the user and item feature matrices as $\mathbf{E}\in \mathbb{R}^{N\times K_1}$ and $\mathbf{F}\in \mathbb{R}^{M \times K_2}$. A toy example of a typical social recommendation setting is illustrated in Figure. \ref{fig:social_rec}. In real-world recommendation systems, the users could hardly interact with all the items, hence the $\mathbf{R}$ matrix is usually very sparse. Here, its entries with question marks are unobserved. 
With these inputs, the problem of social recommendation is formally defined as follows.
\begin{definition}{\textbf{Social Recommendation:}}

\textbf{Given:} User set $\mathcal{U}$, item set $\mathcal{I}$, a social network $G=\{\mathbf{A}, \mathbf{E}\}$, the feature matrix of items $\mathbf{F}$, and the observed user-item interaction $\mathbf{R}$;

\textbf{Output:} The prediction of the interaction scores of the unobserved user-item pairs for users in $\mathcal{U}$.
\end{definition}

\subsection{Preliminaries}

\noindent \textbf{A - Matrix Factorization based Social Recommendation.} The matrix factorization (MF) is originally used on traditional recommender systems where no social relations are available. With social information, the MF methods are extended by using the first-order social relation as regularization \cite{ma2011recommender}. A general formulation could be represented as:
\begin{multline}
    \mathcal{L}_s = \frac{1}{2}\sum_{u=1}^{N}\sum_{i=1}^{M}I_{ui}(\mathbf{R}(u,i)-\mathbf{u}_i\mathbf{v}_j^{\T})^2 + \\ \frac{\beta}{2}\sum_{u=1}^{N}h(\mathbf{u}_i, \{\mathbf{u}_j\}_{j\in\mathcal{N}_i}) 
    +\frac{\lambda_1}{2}||\mathbf{U}||^2_F+\frac{\lambda_2}{2}||\mathbf{V}||^2_F
\end{multline}
where the $I_{ui}$ is an indicator function to indicate whether $(u,i)$ has an observed interaction. $\mathbf{u}_i, \mathbf{v}_j$ are row vectors of $\mathbf{U}$ and $\mathbf{V}$. $h(\mathbf{u}_i, \{\mathbf{u}_j\}_{j\in\mathcal{N}_i})$ is the social regularizer function, in which $\{\mathbf{u}_j\}_{j\in\mathcal{N}_i}$ is a set of user $i$'s neighbors. Representative social regularizer functions are average-based regularization and individual-based regularization, which can be represented as $||\mathbf{u}_i - \frac{\sum_{j\in\mathcal{N}_i}\texttt{sim}(i,j)\cdot\mathbf{u}_j}{\sum_{j\in\mathcal{N}_i}\texttt{sim}(i,j)}||^2_F$, and $\sum_{j\in\mathcal{N}_i}\texttt{sim}(i,j)||\mathbf{u}_i-\mathbf{u}_j||^2_F$, respectively. The $\texttt{sim}(i,j)$ denotes a similarity measure between user $i$ and user $j$.

\noindent \textbf{B - GNN-based Neural Social Recommendation.} As Graph Neural Networks (GNNs) become increasingly popular, numerous GNN-based models have been proposed for social recommendation. Most existing works focus on adopting GNN models for the recommendation task or designing complex GNN layers for the social influence and interest influence diffusion process. The idea is to generate accurate user and item representations. Given the inputs ($G=\{\mathbf{A}, \mathbf{E}\}$, $\mathbf{F}$, $\mathbf{R}$), the user and item representation generated by GNN-based neural model at level $l$ can generally be represented as: 
\begin{subequations}
    \begin{equation}
        \mathbf{u}^{(l)}_i = \Phi(\texttt{GNN}_1(\mathbf{A}, \mathbf{E}), \mathbf{V}^{(l)})
    \end{equation}
    \begin{equation}
        \mathbf{v}^{(l)}_j = \texttt{GNN}_2(\mathbf{R}, \mathbf{E}, \mathbf{F}, \mathbf{U}^{(l)})
    \end{equation}
\end{subequations}
where $\texttt{GNN}_1()$ is used for user representation learning, $\texttt{GNN}_2()$ is used for item representation learning. $\texttt{GNN}_2()$'s inputs include $\mathbf{R}$, which is often treated as an adjacency matrix of user's interest. $\mathbf{U}^{(l)}$, $\mathbf{V}^{(l)}$ are user and item representations at level $l$ respectively, $\Phi()$ is a neural network for leveraging item representation information back to user representation for learning more compatible user/item representations. For example, in \textit{NGCF} \cite{wang2019neural}, the $\Phi()$ function is designed as: $\Phi(\mathbf{e}_i, \mathbf{e}_u) = \textrm{nor}(\mathbf{W}_1\mathbf{e}_i+\mathbf{W}_2(\mathbf{e}_i\odot\mathbf{e}_u))$, where $\textrm{nor}()$ denotes normalization by node and item degrees, $\mathbf{e}_i, \mathbf{e}_u$ denote the item and user representation at one level of GNN layers, and $\mathbf{W}_1, \mathbf{W}_2$ denote the learnable weights. Generally, this direction of methods does not sufficiently study the specific strategy of negative sampling, how to design GNN layers for mitigating oversmoothing, or how to utilize user-item interaction for message aggregation. As more recent works focus on designing complicate structures for $\texttt{GNN}_1()$, $\texttt{GNN}_2()$ and $\Phi()$, our findings suggest that instead of stacking sophisticated message-passing layers, a succinct GNN can still achieve superior performance with carefully designed negative sampling strategy, initialization, and interest propagation.


\section{Proposed Model}
\label{sec:method}
In this section, we present the proposed model with the generative negative sampling strategy, followed by some analysis. 

\subsection{Model Architecture}
\begin{figure*}
    \centering
    \includegraphics[width=0.95\textwidth, height=0.36\textwidth]{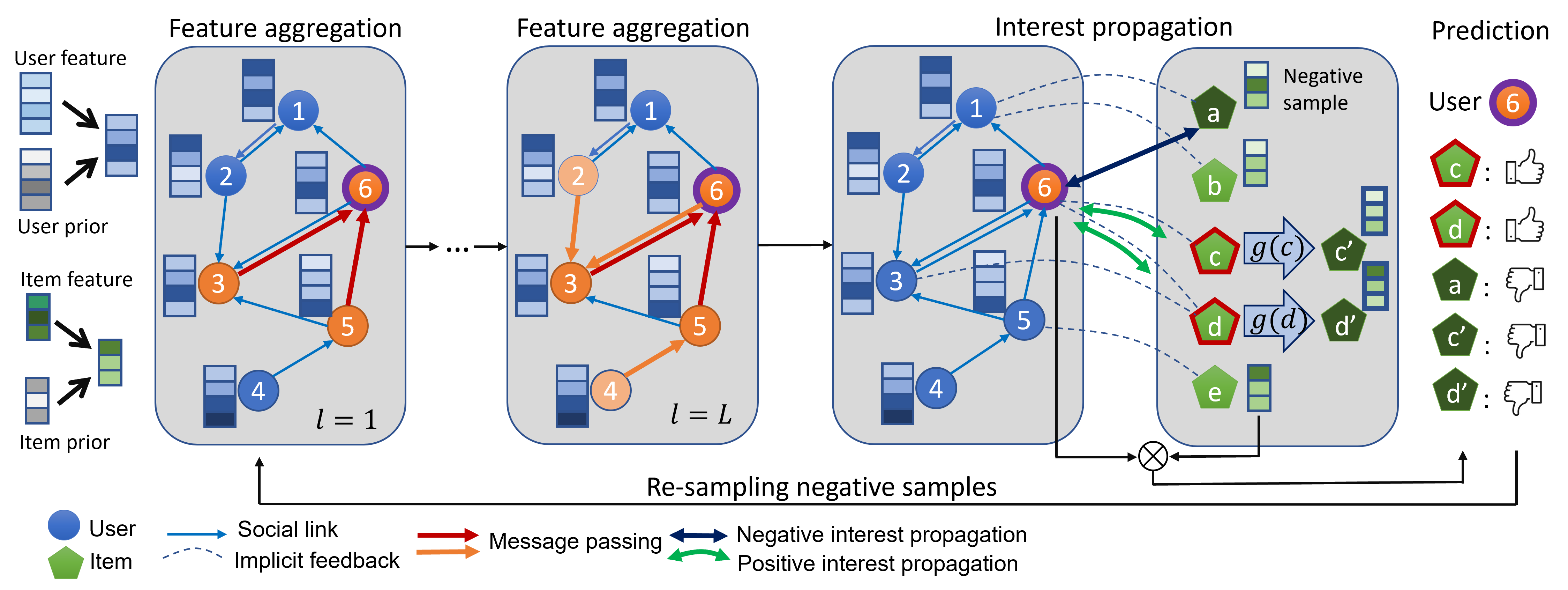}
    \caption{The overall model architecture of \nemo. Best viewed in color.}
    \label{fig:model}
\end{figure*}

We first show the succinct multi-network GNN-based neural model architecture. The key ideas of the proposed model are two-fold. First, the user-user social relations are utilized only for feature aggregation among users, without the complicated architecture designs from various existing GNN models. Here, we hypothesis that the major benefit of the GNN models for social recommendation originates from the local feature aggregation among users in the social network. This idea aligns with the observations made by traditional recommendation techniques as well \cite{ma2011recommender} Furthermore, GNN models often suffer from oversmoothing with multiple layers (\cite{chen2020measuring}). As we will see from the empirical evaluation, users that are multiple hops away provide little or even negative impact to the users' representation learning. Thus the common multi-layer GNN architecture and the nonlinearity between GNN layers can be refrained, which also reduces the model complexity. Second, apart from the user-user graph, we combine both the positive and the sampled negative user-item interactions into a heterogeneous bipartite graph, which is then used for interest propagation across items and users. The idea is that the local item features of a user consist of the interest context of the user, meanwhile the local user features of an item consist of the property context of the item. Through message passing on a heterogeneous graph, the item interest of a user could be propagated to the neighboring users, and the product property of an item could also be propagated to its neighboring items. As a result, the learned user and item representation could be more compatible when used for recommendation/link inference. The model architecture is illustrated in Figure \ref{fig:model}. 

Figure \ref{fig:model} also shows how the input user/item features are processed within one epoch. At the beginning of each epoch, for each user, we uniformly sample a fixed number of items, which do not have interactions with the user. We temporarily treat these items as `negative' items for interest propagation in the model. Note that the `negative' items are not fixed for each epoch. If there is already a set of `negative' samples for each user from the last epoch, a new set of `negative' samples will be re-sampled. A detailed negative sampling strategy is elaborated in Section \ref{sec:negative}. 

In the model, first of all, the user/item features are fed into a multi-layer perceptron (MLP), and then combined with a user/item prior embedding, which is also known as the `free embedding' in some existing works (\cite{wu2020diffnet++}, \cite{koren2008factorization}), in order to obtain user/item hidden representations for further processing. For example, given user $u$ and item $i$, the hidden representations can be represented as:

\begin{subequations}
    \begin{equation}\label{h1}
    \mathbf{H}_1(u,:) = \texttt{MLP}_1(\mathbf{E}(u,:)) + \mathbf{P}(u,:)
    \end{equation}
    \begin{equation}\label{h2}
    \mathbf{H}_2(i,:) = \texttt{MLP}_2(\mathbf{F}(i,:)) + \mathbf{Q}(i,:)
    \end{equation}
\end{subequations}
where $\mathbf{P}$ and $\mathbf{Q}$ denote the learnable prior representations of users and items, whose initialization $\mathbf{P}(u,:), \mathbf{Q}(i,:) \sim \mathcal{N}(\mathbf{0}, \gamma^2\mathbf{I})$ follows a Gaussian distribution with zero mean and standard deviation $\gamma>0$. The intuition of imposing a prior with Gaussian distribution originates from the classic probabilistic matrix factorization for traditional recommender systems. \cite{mnih2007probabilistic} shows that with Gaussian noise as the prior distribution of the latent representations, maximizing the log-posterior over user and item hidden representation is equivalent to minimizing a squared error objective with quadratic regularization. The prior in Eq. \eqref{h1} \eqref{h2} work in a similar fashion (see analysis in Section \ref{sec:discussion}).

Second, user hidden representation is used for the $L-$level feature aggregation within the social network. In each level, the hidden representations are passed from a source node to a target node of each edge, and the representations are summed up as the updated user hidden representation ($\mathbf{H}_1$ and $\mathbf{H}_2$). Third, the resulting user representations and the item representations are diffused in two directions by users' interactions with both positive and the sampled negative items, for positive and negative interest propagation. Here, the users and positive/negative items form a heterogeneous bipartite graph $G_{b}$. We use two different MLPs for modeling the positive and negative interest propagation. Then, the positive items are fed into a neural module for generating negative samples. Suppose that $\mathbf{A}^{+}_{b}\in\mathbb{R}^{N\times M}$ denotes the adjacency matrix induced from positive interactions of $G_{b}$. $\mathbf{A}^{+}_{b}(i,j)=1$ if user $i$ has iteractions with item $j$. $\mathbf{A}^{-}_{b}\in \mathbb{R}^{N\times M}$ denotes the adjacency matrix induced from sampled negative interactions of $G_{b}$. $\mathbf{A}^{-}_{b}(i,j)=1$ if user $i$ and item $j$ are sampled as negative pairs. Generally, the intuition behind this module is to generate compatible user (item) embeddings with the assist of the information from item (user).  The user embedding $\mathbf{\hat{H}}_1$ and item embedding $\mathbf{\hat{H}}_2$ after interest propagation are:
\begin{equation}
    \mathbf{\hat{H}}_1 = \mathbf{A}^{+}_b \cdot \mathbf{H}_2 \cdot (\mathbf{D}^{+}_u)^{-1} + \mathbf{H}_1
\end{equation}
\begin{multline}
    \mathbf{\hat{H}}_2 = \texttt{MLP}_p((\mathbf{A}^{+}_b)^{\T} \cdot \mathbf{H}_1 \cdot (\mathbf{D}^{+}_i)^{-1})   \\ + \texttt{MLP}_n((\mathbf{A}^{-}_b)^{\T} \cdot  \mathbf{H}_1 \cdot (\mathbf{D}^{-}_i)^{-1}) + \mathbf{H}_2
\end{multline}
where $\mathbf{D}^{+}_u, \mathbf{D}^{+}_i, \mathbf{D}^{-}_i$ are diagonal degree matrices for representation normalization. $\mathbf{D}^{+}_u(x, x)=\sum_{y}\mathbf{A}^{+}_b(x, y)$, $\mathbf{D}^{+}_i(x, x)=\sum_{x}\mathbf{A}^{+}_b(x, y)$ and $\mathbf{D}^{-}_u(x, x)=\sum_{y}\mathbf{A}^{-}_b(x, y)$. $\texttt{MLP}_p()$ and $\texttt{MLP}_n()$ are two MLP modules for the users' positive and negative interest propagation in item representation respectively.
Finally, after the interest propagation, the representations of users and items are used for calculating the final predictions by inner product. The details for loss function is discussed in Section \ref{sec:train}.

\subsection{Generative Negative Sampling}\label{sec:negative}
Here, we present the proposed generative negative sampling strategy for the social recommendation problem.

Numerous existing works adopt uniform negative sampling (UNS) from the unobserved items. However, it is unreasonable to naively assume that the unobserved items are equal to negative samples, because it might introduce bias into the negative samples, and the UNS method is not able to generate hard negative examples. In order to alleviate this issue, inspired by recent studies on mixup (\cite{yu2021mixup}, \cite{huang2021mixgcf}), which linearly interpolates pairs of examples for data augmentation in various tasks, we introduce the generative negative sampling. Our key idea is to generate negative samples through neural networks in the continuous embedding space of items. For a given user $u$, the neural generator takes the hidden representations of true positive items $\mathbf{z}_i$ as inputs, and the representations of uniformly sampled unobserved items $\{i_1, ..., i_p\}$ as offsets:
\begin{equation}\label{eq:z}
    \mathbf{z'}_i = \alpha \cdot g(\mathbf{z}_i) + (1-\alpha) \cdot \pooling_{i_1,...,i_p\in \Omega_{-}}({\mathbf{z}_{i_1}, ..., \mathbf{z}_{i_p}})
\end{equation}
\hh{i think this could be one of most interesting contribution of the paper. can we give any (theoretic) analysis on why Eq6 is a good choice?}\bx{working on this}where $g(\mathbf{z}_i) = \sigma(\mathbf{W}_2 \cdot \sigma(\mathbf{W}_1 \cdot \mathbf{z}_i + \mathbf{b}_1) + \mathbf{b}_2)$ is a MLP, with $\mathbf{W}_1, \mathbf{W}_2, \mathbf{b}_1, \mathbf{b}_2$ as learnable parameters, and $\sigma()$ is a Sigmoid function. $\Omega_{-}$ is the set of sampled unobserved items. $\pooling()$ is the Pooling function which transforms the input representations of the sampled unobserved items into a combined representation. $\alpha$ is a weighting scalar for the first MLP term and the second offset term of unobserved samples.
The intuition of the first MLP term is to deviate the true item representation from the embedding space in order to generate a fake item representation, whose corresponding item might not exist. The intuition of the second term is to bump such deviation in the direction of unobserved samples, in which the true negative samples might exist. The above negative sample generator is learned with the model in an end-to-end fashion. 

There are two major advantages by the generative negative sampling. First, the generated samples are mixed with information from positive items, resulting in harder negative samples compared with samples by UNS. Second, the generated fake samples can be combined with the real samples as an augmentation for the dataset. In this way, the proposed model is potentially more generalizable because of the interpolation of embedding spaces with augmentation during training.

\subsection{Implementation Details}
Here, we further present some implementation details. 

\noindent \textbf{A - Choice of $\alpha$.} As suggested by \cite{yu2021mixup}, the selection of $\alpha$ affects the model generalization, and it is often solved by sampling from a distribution. Here we use \bx{TODO:check}Beta distribution due to its strong empirical performance. 

\noindent \textbf{B - Static Negative Re-sampling.}
As a natural extension for UNS, we can uniformly re-sample the negative samples from unobserved items in every epoch of the training\hh{why do we call it 'static'?}, instead of using one fixed negative item set. Static means the sampling method does not consider the model output dynamically as in dynamic negative sampling. Compared with fixing the negative item set, this could potentially prevent overfitting and also improve the model's generalization (see Section \ref{sec:experiments}).

\noindent \textbf{C - Dynamic Negative Re-sampling.} 
We also adopt dynamic negative sampling for mitigating the limitations of UNS and obtaining hard negative samples dynamically at each epoch. Specifically, the unobserved items are ranked by the model's output rating scores at each training epoch. Then the negative samples are extracted from the top ranked items, which are supposed to be the hard examples.

\subsection{Training}\label{sec:train}
Instead of leveraging the commonly used Bayesian Personalized Ranking (BPR) loss, we adopt the simple Mean Squared Error (MSE) loss, which is easier to implement for our model and meanwhile with superior performance. We briefly discuss two limitations of the BPR loss in terms of practical implementation. First, given a pair of positive and negative samples, BPR loss ranks the positive sample higher than the negative sample. If the pair is uniformly sampled from the observed and unobserved items, it always makes the ratio of positive and negative items equal to $1:1$, which restricts the model generalization. Second, for $C$ positive items, we sample $rC$ negative samples using negative sampling ratio $r$, and consider every combination of positive and negative item pairs in BPR loss. The computational complexity is $O(N\cdot rC^2)$, which is significantly larger than MSE loss ($O(N\cdot (C+rC))$). Our loss function is given as follows.
\begin{equation}
    \mathcal{L} = \sum_{(u,i)\in\Omega_{+}\cup\Omega_{-}} ||\mathbf{R}(u,i)-\mathbf{\hat{H}}_1(u,:)\mathbf{\hat{H}}_2(i,:)^{\T}||^2_2
\end{equation}
The predicted rating of $(u,i)$ is calculated as the inner product of representation vectors $\mathbf{\hat{H}}_1(u,:)\mathbf{\hat{H}}_2(i,:)^{\T}$. $\Omega_{+}$ and $\Omega_{-}$ are positive and negative user-item pair set respectively. Here, for $\Omega_{-}$, we combine the uniformly re-sampled items with the generated fake items as a given user's final negative samples to obtain the best performance. We adopt the Adam optimizer which shows more stable convergence than other optimizers. The Adam optimizer is applied with a weight decay of $0.001$. 

\subsection{Analysis}\label{sec:discussion}
\bx{1. prior's effect 2.complexity}
In this subsection, we first discuss the effect of the Gaussian prior, and then give the complexity analysis for \nemo.\hh{can we provide any analysis of Eq6, say based on Jian Tang's comment?}

\noindent \textbf{A - Gaussian Prior.} Here, under the same assumption as probabilistic matrix factorization, the conditional probability of the observed user-item interaction follows a Gaussian distribution:
\begin{equation}\label{eq:p_R}
    p(\mathbf{R}|\mathbf{P}, \mathbf{Q}, \gamma_R^2) = \prod^N_{u=1}\prod^M_{i=1} [\mathcal{N}(\mathbf{R}(u,i)|f(\mathbf{p}_u)f(\mathbf{q}_i)^{\T}, \gamma_r^2)]^{I_{u,i}^{\mathbf{R}}}
\end{equation}
where $\mathbf{p}_u$ and $\mathbf{q}_i$ are the $u$-th row vector and the $i$-th row vector in $\mathbf{P}, \mathbf{Q}$ respectively. $f()$ function represents the model, and here we assume that the rest of the parameters are fixed. $I_{u,i}^{\mathbf{R}}$ is an indicator function, which is equal to $1$ if $(u,i)$ has observed interactions, and otherwise $0$. Given that $\mathbf{P}(u,:), \mathbf{Q}(i,:) \sim \mathcal{N}(\mathbf{0}, \gamma^2\mathbf{I})$, the posterior probability of the $\mathbf{Q}$ and $\mathbf{P}$ is:
\begin{equation}\label{eq:p_PQ}
    p(\mathbf{P, Q}|\mathbf{R}, \mathbf{A}, \gamma^2, \gamma^2_R) \propto p(\mathbf{R}|\mathbf{P}, \mathbf{Q}, \gamma_R^2)p(\mathbf{P}|\gamma^2)p(\mathbf{Q}|\gamma^2)
\end{equation}
We plug Eq. \eqref{eq:p_R}, $p(\mathbf{P}|\gamma^2)$, and $p(\mathbf{Q}|\gamma^2)$ into Eq. \eqref{eq:p_PQ}, and maximize the log-posterior becomes equal to minimizing the following formula:
\begin{multline}\label{eq:L'}
    \mathcal{L'} = \frac{1}{2}\sum_{u=1}^{N}\sum_{i=1}^{M} I_{u,i}^{\mathbf{R}}(\mathbf{R}(u,i)-f(\mathbf{p}_u)f(\mathbf{q}_i)^{\T}) + \\ \frac{\gamma_R^2}{2\gamma^2}\sum_{u=1}^{N}||\mathbf{P}(u,:)||^2_2 + \frac{\gamma_R^2}{2\gamma^2}\sum_{i=1}^{M}||\mathbf{Q}(i,:)||^2_2
\end{multline}
We can see that the last two terms in Eq. \eqref{eq:L'} indicate the L2-regularization of the learnable prior representations $\mathbf{P,Q}$.

\noindent \textbf{B - Complexity Analysis.} The major computational hurdle lies in the user feature aggregation and the interest propagation. Suppose that the adjacency matrix of user-user social network contains $m_1$ non-zero entries, and the adjacency matrix of the bipartite user-item graph contains $m_2$ and $m_3$ non-zero entries for $\mathbf{A}_b^{+}, \mathbf{A}_b^{-}$ respectively. Let the dimension of the user/item hidden representations equal to $d$. Then the time complexity of the proposed model is $O(iter \cdot (K_1m_1+d(m_2+m_3)))$, where $iter$ indicates the number of iterations in all forward pass. Since the adjacency matrix $\mathbf{A}$, $\mathbf{A}_b^{+}$, and $\mathbf{A}_b^{-}$ are all usually sparse in real-world data, $m_1, m_2, m_3$ are usually comparable or smaller than \hh{or can we say $m_1, m_2, m_3 \ll M, N$}the magnitude of $N, M$, and are much smaller when compared with the magnitude of $N^2$ and $M^2$.

\section{Experiments}
\label{sec:experiments}
In this section, we present the experimental results on real-world datasets to show the effectiveness of \nemo. 

\subsection{Experimental Setting}
We use two widely used benchmark datasets in the experiments, and their statistics are shown in Table. \ref{tab:dataset}.
\begin{table}[h]
    \centering
    \caption{The statistics of the two benchmark datasets.}
    \label{tab:dataset}
    \begin{tabular}{|c|c|c|}
    \hline
         Dataset & Yelp & Flickr   \\
         \hline
         \# of users & 17,237 & 8,358 \\
         \hline
         \# of items & 38,342 & 82,120 \\
         \hline
         \# of ratings & 204,448 & 143,765 \\
         \hline
         \# of observed links & 0.03\% & 0.05\% \\
         \hline
         Link density & 0.05\% & 0.27\% \\ 
         \hline
    \end{tabular}
\end{table}

\noindent \textbf{A - Datasets and Pre-processing.} \url{Yelp.com} is an online website that publishes crowd-sourced reviews and ratings about businesses. Users can also connect to each other to have a social relation through Yelp. The ratings are in the range of $[0,5]$, and the reviews are usually text and images. The \textit{Yelp} dataset\footnote{https://www.yelp.com/dataset} mainly consists of information about users, businesses, and reviews. For the pre-processing, we use the same setting as the baselines \cite{wu2019neural}, \cite{wu2020diffnet++}. The user-user social network is constructed via the user's friend information. The user/item feature vectors are calculated by averaging all the learned word embeddings of the user/item via the Word2vec model. 

\textit{Flickr} is an American image hosting and video hosting service, as well as an online community. Users can follow each other and build social relations. The items (photos and videos) can be upvoted by users as implicit feedback. The \textit{Flickr} dataset is shared by \cite{wu2019neural}. For the pre-processing, the user feature vectors are calculated by averaging the image feature representations s/he liked. The image representations are generated \hh{this sentence seems broken. check} via a VGG16 convolutional neural network. 

\noindent \textbf{B - Baseline Methods.} We compare the proposed model with three types of baseline methods: (1) one representative traditional recommendation method without the usage of social relations (\textit{BPR} \cite{rendle2012bpr}); (2) one representative traditional social recommendation methods which models the first-order social relations (\textit{CNSR} \cite{wu2018collaborative}); and (3) six state-of-the-art GNN-based social recommendation methods (\textit{GraphRec} \cite{fan2019graph}, \textit{PinSage} \cite{ying2018graph}, \textit{NGCF} \cite{wang2019neural}, \textit{DiffNet} \cite{wu2019neural}, \textit{DiffNet++} \cite{wu2020diffnet++}, \textit{DiffNetLG} \cite{song2021social})\footnote{The results of \textit{DiffNetLG} are from the report of the paper which only uses \textit{Yelp} dataset due to no open-sourced code.}.

\noindent \textbf{C - Experimental Settings.}
For the metrics, we use two most commonly used metrics for recommendation, Hit Ratio (HR@$K$) and Normalized Discounted Cumulative Gain (NDCG@$K$), where $K\in\{5, 10, 15\}$. In the experiment, the training, validation, and testing data ratio is equal to $8:1:1$, and we use the exact same split for all baselines. In the evaluation, we sample $1,000$ unrated items for each user and combine them with the rated items for the calculation of HR@$K$ and NDCG@$K$. The results are averaged over five runs. 

\noindent \textbf{D - Hyperparameter Settings.} For \textit{Yelp} dataset, the number of uniformly sampled unobserved items for each user is 8. For \textit{Flickr} dataset, the number of uniformly sampled unobserved items for each user is 15. 
For both datasets, we use 2 layers of user feature aggregation, 200 as batch size, and 0.001 as learning rate. 

\subsection{Effectiveness Results}
\begin{table}[h]
\caption{Social recommendation comparison on \textit{Yelp} dataset.}
\fontsize{7}{9}\selectfont
    \centering
    \resizebox{8.5cm}{!} 
    {
    \begin{tabular}{l|c|c|c||c|c|c}
    \hline
            & \multicolumn{3}{c}{HR@K} &  \multicolumn{3}{c}{NDCG@K} \\ \hline
         Models & K=5 & K=10 & K=15 & K=5 & K=10 & K=15 \\ \hline
         
         BPR & 0.1695 & 0.2632 & 0.3252 & 0.1231 & 0.1554 &  0.1758 \\  \hline
         
         CNSR & 0.1877 & 0.2904 & 0.3458 & 0.1389 & 0.1746 & 0.1912  \\  \hline
         
         GraphRec & 0.1915 & 0.2912 & 0.3623 & 0.1279 & 0.1812 & 0.1956  \\ 
         
         PinSage & 0.2105 & 0.3049 & 0.3863 & 0.1539 & 0.1828 & 0.2130  \\ 
         
         NGCF & 0.1992 & 0.3042 & 0.3863 & 0.1450 & 0.1828 & 0.2041 \\
         
         DiffNet & 0.2276 & 0.3461 & 0.4217 & 0.1679 & 0.2118 & 0.2307  \\ 
         
         DiffNet++ & 0.2503  & 0.3694 & \textbf{0.4493} & 0.1841 & 0.2263 & 0.2497  \\
         
         DiffNetLG & 0.2599 & 0.3711 & \underline{0.4473} & 0.1941 & 0.2333 & 0.2586  \\ \hline
         
         \rowcolor{Gray}
         Ours (Stat.) & \underline{0.3857} & \underline{0.3918} & 0.3982 & \underline{0.4093} & \underline{0.4105} & \underline{0.4122}  \\ 
         
         \rowcolor{Gray}
         Ours (Gen.) & \textbf{0.3956} & \textbf{0.4025} & 0.4146 & \textbf{0.4198} & \textbf{0.4212} & \textbf{0.4231}  \\
         \hline
         
    \end{tabular}
    }
    \label{tab:yelp}
\end{table}
The performance comparison is shown in Tables \ref{tab:yelp} and  \ref{tab:flickr}. `Stat.' means training with static re-sampling strategy, and `Gen.' indicates training with generative negative sampling strategy. The best performances are shown in bold font and the second-best performances are shown with underlines. On \textit{Yelp} dataset, except for HR@15, our proposed model significantly outperforms all baseline methods. Specifically, the HR@5 is improved by 34.3\% compared with the best baseline. The NDCG@$K$ is also significantly improved. For example, the NDCG@15 is increased by up to 38.8\%. On \textit{Flickr} dataset, similar observations can be made. The HR@5 is increased by 8.7\% and the NDCG@15 is increased by 13.3\% compared with the best baseline. Furthermore, the proposed model shows even more improvement for small $K$.
         
         
         
         
         
         
         
         
         
         
         

\begin{table}
\caption{Social recommendation comparison on \textit{Flickr} dataset.\hh{do we have any potential reason why we are significantly worse than diffnet+ for HR@15}}
\fontsize{7}{9}\selectfont
    \centering
    \resizebox{8.6cm}{!} 
    {
    \begin{tabular}{l|c|c|c||c|c|c}
    \hline
            & \multicolumn{3}{c}{HR@K} &  \multicolumn{3}{c}{NDCG@K} \\ \hline
         Models & K=5 & K=10 & K=15 & K=5 & K=10 & K=15 \\ \hline
         
         BPR & 0.0651 & 0.0795 & 0.1037 & 0.0603 & 0.0628 &  0.0732 \\  \hline
         
         CNSR & 0.0920 & 0.1229 & 0.1445 & 0.0791 & 0.0978 & 0.1057  \\  \hline
         
         GraphRec & 0.0931 & 0.1231 & 0.1482 & 0.0784 & 0.0930 & 0.0992  \\ 
         
         PinSage & 0.0934 & 0.1257 & 0.1502 & 0.0844 & 0.0998 & 0.1046  \\ 
         
         NGCF & 0.0891 & 0.1189 & 0.1399 & 0.0819 & 0.0945 & 0.0998 \\
         
         DiffNet & 0.1178 & 0.1657 & 0.1855 & 0.1072 & 0.1271 & 0.1301  \\ 
         
         DiffNet++ & 0.1412  & \textbf{0.1832} & \textbf{0.2203} & 0.1296 & 0.1420 & 0.1544  \\ \hline
         
         \rowcolor{Gray}
         Ours (Stat.) & \underline{0.1452} & 0.1411 & 0.1412 & \underline{0.1725} & \underline{0.1674} & \underline{0.1667}  \\ 
         
         \rowcolor{Gray}
         Ours (Gen.) & \textbf{0.1547} & \underline{0.1504} & \underline{0.1469} & \textbf{0.1850} & \textbf{0.1792} & \textbf{0.1781}  \\
         \hline
         
    \end{tabular}
    }
    \label{tab:flickr}
\end{table}

\subsection{Ablation Study}
\setlength{\columnsep}{13pt}%
\begin{wrapfigure}{R}{0.25\textwidth}
\centering
\includegraphics[width=0.25\textwidth]{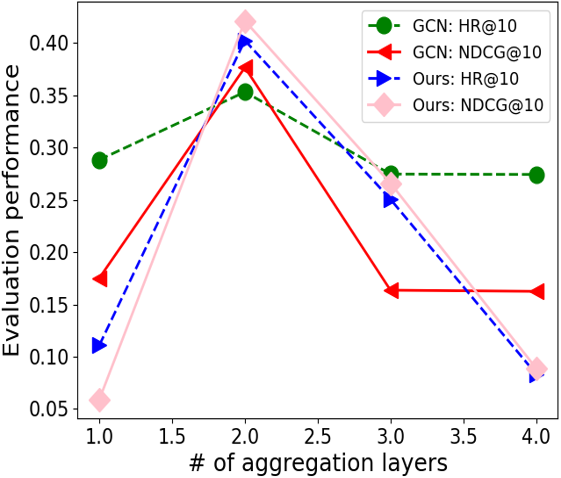}
\caption{\label{fig:sensitivity}The impact of aggregation layers}
\end{wrapfigure} 
The ablation study results are shown in Tables \ref{tab:ab_yelp} and  \ref{tab:ab_flickr}. The `Uni. prior' represents the model trained with uniformly distributed prior user/item representations. The `No prior' represents training without the prior user/item representation. `GCN' denotes our model variant in which the user feature aggregation is replaced with a Graph Convolution Network. `No re-sample' denotes training without the negative re-sampling process at each epoch. As we can see from the results, firstly the prior representation has a huge impact on the model performance. Without such prior, the performance could drop up to 59.8\% for HR. Initializing the prior with proper distribution is also important since Gaussian distribution outperforms the uniform distribution. As we discuss in Section \ref{sec:discussion}, the Gaussian distribution regularizes the learnable prior representations. Secondly, using the GNN model does not positively contribute to the final performance. As we discuss in Section \ref{sec:method}, the non-linearity and complex design in GNN models might have little or even negative influence. Instead, using succinct two-layer linear aggregation performs the best. Thirdly, the re-sampling strategy is crucial and could improve the model's ability of generalization. Generally, the generative negative sampling outperforms the static negative sampling and further outperforms no re-sampling. 

As discussed in Section \ref{sec:method}, we show the impact of aggregation layers for user features in Figure \ref{fig:sensitivity}. Both the proposed model and the model using GCN module are tested. As we can see, for both model variants, the performance reaches the peak at 2 layers of aggregation and quickly drops at a larger number of layers. It also suggests that it might not be necessary to use deep GNN models in many social recommendation tasks as the representation becomes oversmoothed with large layers. 

\begin{table}
\caption{Results for ablation study on \textit{Yelp} dataset.}
\fontsize{7}{9}\selectfont
    \centering
    \resizebox{8.6cm}{!} 
    {
    \begin{tabular}{l|c|c|c||c|c|c}
    \hline
            & \multicolumn{3}{c}{HR@K} &  \multicolumn{3}{c}{NDCG@K} \\ \hline
         Models & K=5 & K=10 & K=15 & K=5 & K=10 & K=15 \\ \hline
         
         Uni. prior & 0.2718 & 0.2747 & 0.2790 & 0.2935 & 0.2937 &  0.2948 \\  
         
         No prior & 0.1591 & 0.1617 & 0.1654 & 0.1698 & 0.1703 & 0.1714  \\  
         
         GCN & 0.3541 & 0.3532 & 0.3533 & 0.3777 & 0.3767 & 0.3766  \\ 
         
         No re-sample & 0.3713 & 0.3764 & 0.3821 & 0.3950 & 0.3967 & 0.3983  \\ 

         Ours (Stat.) & \underline{0.3857} & \underline{0.3918} & \underline{0.3982} & \underline{0.4093} & \underline{0.4105} & \underline{0.4122}  \\ 
         
         \rowcolor{Gray}
         Ours (Gen.) & \textbf{0.3956} & \textbf{0.4025} & \textbf{0.4146} & \textbf{0.4198} & \textbf{0.4212} & \textbf{0.4231}  \\
         \hline
         
    \end{tabular}
    }
    \label{tab:ab_yelp}
\end{table}

\begin{table}
\caption{Results for ablation study on \textit{Flickr} dataset.}
\fontsize{7}{9}\selectfont
    \centering
    \resizebox{8.6cm}{!} 
    {
    \begin{tabular}{l|c|c|c||c|c|c}
    \hline
            & \multicolumn{3}{c}{HR@K} &  \multicolumn{3}{c}{NDCG@K} \\ \hline
         Models & K=5 & K=10 & K=15 & K=5 & K=10 & K=15 \\ \hline
         
         Uni. prior & \underline{0.1479} & \underline{0.1442} & \underline{0.1443} & \underline{0.1758} & \underline{0.1706} &  \underline{0.1700} \\ 
         
         No prior & 0.0840 & 0.0812 & 0.0808 & 0.0991 & 0.0957 & 0.0951  \\  
         
         GCN & 0.0687 & 0.0890 & 0.1125 & 0.0693 & 0.0687 & 0.0721  \\ 
         
         No re-sample & 0.1045 & 0.1022 & 0.1020 & 0.1213 & 0.1186 & 0.1182  \\  
         
         Ours (Stat.) & 0.1452 & 0.1411 & 0.1412 & 0.1725 & 0.1674 & 0.1667  \\ 
         
         \rowcolor{Gray}
         Ours (Gen.) & \textbf{0.1547} & \textbf{0.1504} & \textbf{0.1469} & \textbf{0.1850} & \textbf{0.1792} & \textbf{0.1781}  \\
         \hline
         
    \end{tabular}
    }
    \label{tab:ab_flickr}
\end{table}

\vspace{-0.6\baselineskip}
\section{Related work}
\label{sec:related-work}
The related works can be divided into the classical social recommendation methods, and the recent GNN models.

\noindent \textbf{A - Classical Social Recommendation Methods.} Generally speaking, there are two definitions of social recommendation. According to \cite{tang2013social}, the narrow definition says that the social recommendation is any recommendation with online social relations as an additional input. The broad definition of social recommendation is any recommender systems that target at social media domains, such as recommending users, tags, communities, etc. The majority of the classical social recommendation methods focus on the narrow definition. To name a few representative works, \textit{SocialMF} \cite{jamali2010matrix} incorporates the mechanism of trust propagation into the matrix factorization-based model, and shows its effectiveness in tackling cold-start problems. \textit{TrustSVD} \cite{guo2015trustsvd} extends the idea of trust propagation to leverage both explicit and implicit influence of ratings and trust in the matrix factorization model. \textit{ContexMF} \cite{jiang2014scalable} proposes a probabilistic matrix factorization method to fuse the individual preference and interpersonal influence. \textit{CNSR} \cite{wu2018collaborative} is one of the earliest works to introduce neural models in social recommendation. Yao et al. propose a dual-regularized model \cite{yao2014dual} for one-class collaborative filtering, which tackles the sparseness challenge in the problem by exploiting the side information from both users and items. FASCINATE \cite{chen2016fascinate} by Chen et al. develop a multi-layered graph-based approach for cross-Layer dependency inference, which can be seen as a generalization of the underlying problem of social recommendation by graph-based solution. Similarly, \cite{du2018fasten} and \cite{du2021sylvester} by Du et al. are general Sylvester equation-based methods for across-network node association inference, which can also be regarded as generalization of social recommendation problem. \hh{maybe after the paper is accepted, cite Yuan's wi\_Zan from CIKM12 and cc's KDD16 paper}
It proposes two modules, namely a social embedding and a collaborative neural recommendation part, and further combines them in a joint learning framework.

\noindent \textbf{B - GNN-based Models.} Numerous GNN-based models have been proposed. We review some of the most relevant works here. \textit{PinSage} \cite{ying2018graph} is one of the earliest works to apply GNNs on recommender systems. The model constructs convolution operations via random walks to generate embeddings of items which incorporate both graph topology and feature information. \textit{NGCF} \cite{wang2019neural} proposes to propagate the user/item embeddings on the user-item bipartite graph for modeling the high-order connectivity. \textit{GraphRec} \cite{fan2019graph} proposes a neural model to jointly capture the interactions and opinions in the user-item graph, in order to handle the heterogeneity issue in the user-user and user-item relation. Recently, \textit{DiffNet} \cite{wu2019neural} and \textit{DiffNet++} \cite{wu2020diffnet++} propose a GNN-based model to combine the social influence diffusion with the interest diffusion, with different designs of GNN architectures. As a follow-up, \textit{DiffNetLG} \cite{song2021social} models both local
implicit influence of users on unobserved interpersonal relations, and global implicit influence of items broadcasted to users. \textit{LightGCN} \cite{he2020lightgcn} simplifies the GCN model in recommendation to only keep the neighbor aggregation for collaborative filtering. \textit{RecQ} \cite{yu2021self} develops a hypergraph-based model to model the high-order user relations in social recommendation, which is one of the promising directions in this line of research. A related work by Du et al. \cite{du2021hypergraph, du2022self} suggests that the pre-training on hypergraphs could help to further improve the representation learning in various downstream tasks. Zhang et al. propose a subgraph-based GNN model for bundle recommendation that shows the effectiveness of subgraph embedding in this task. 

\section{Conclusion}
\label{sec:conclusion}
In this paper, we propose a succinct GNN model with focused user feature aggregation and interest propagation. Furthermore, we leverage both positive and negative samples for users' preference diffusion between the representations of users and items in order to learn more compatible embeddings. Lastly, we propose a generative negative sampling approach to interpolate hard negative samples for improving the model's ability of generalization. Empirical results show that the proposed model significantly outperforms the state-of-the-art GNN-based models.  
\bibliographystyle{siam}
\bibliography{reference.bib}

\end{document}